# Content-Based Filtering for Video Sharing Social Networks

Eduardo Valle[*], Sandra de Avila[†], Antonio da Luz Jr.[†‡],
Fillipe de Souza[†], Marcelo Coelho[†§], Arnaldo Araújo[†]

[*] *RECOD Lab — IC / UNICAMP, Campinas, SP, Brazil*
[†] *NPDI Lab — DCC / UFMG, Belo Horizonte, MG, Brazil*
[‡] *Tocantins Federal Institute of Education and Technology — IFTO, Paraíso, TO, Brazil*
[§] *Preparatory School of Air Cadets — EPCAR, Barbacena, MG, Brazil*

E-mails: *mail@eduardovalle.com, {sandra, daluz, fdms, mcoelho, arnaldo}@dcc.ufmg.br*

Corresponding Author: Eduardo Valle, mail@eduardovalle.com
phone: +19 3521-5887, fax: +19 3521-5847

## Abstract

*In this paper we compare the use of several features in the task of content filtering for video social networks, a very challenging task, not only because the unwanted content is related to very high-level semantic concepts (e.g., pornography, violence, etc.) but also because videos from social networks are extremely assorted, preventing the use of constrained* a priori *information. We propose a simple method, able to combine diverse evidence, coming from different features and various video elements (entire video, shots, frames, keyframes, etc.). We evaluate our method in three social network applications, related to the detection of unwanted content — pornographic videos, violent videos, and videos posted to artificially manipulate popularity scores. Using challenging test databases, we show that this simple scheme is able to obtain good results, provided that adequate features are chosen. Moreover, we establish a representation using codebooks of spatiotemporal local descriptors as critical to the success of the method in all three contexts. This is consequential, since the state-of-the-art still relies heavily on static features for the tasks addressed.*

## Keywords

Content filtering; Video classification; Local features; Bags of Visual Features; Spatiotemporal features; Majority voting; Pornography detection; Violence detection, Popularity evaluation; Video social networks.

# 1 Introduction

Content-based classification and retrieval of visual documents by high-level semantic concepts has been an elusive goal pursued by the scientific community for the last 20 years. The persistent absence of a general solution attests the task difficulty, which is in great part brought by the much discussed "semantic gap" between the low-level representation of the data (pixels, frames, etc.) and the high-level concepts one wants to take into account.

We have, however, witnessed many important breakthroughs. In what concerns the description of visual documents, we have watched not only the inception and evolution of local features [24], but mainly the introduction of representations based on codebooks [5], which have allowed to conciliate the discriminating power of the former with the generalization abilities required by high-level semantic tasks. Specifically for video, the introduction of "motion-aware" local features, which take into account the dual nature of that media, at the same time spatial *and* temporal, has been an important achievement [6][13][14][15][16]. Meanwhile, the development of machine learning algorithms, like SVM [23] has created an effective framework for complex classification tasks.

In this paper, we are concerned with the detection of unwanted content on video sharing social networks — online communities built upon the production, sharing and watching of short video clips, which have been nourished by the popularization of broadband web access and the availability of cheap video acquisition devices. The crowds of users who employ the services of websites like *Dailymotion*, *MetaCafe* and *YouTube*, not only post and watch videos, but also share ratings, comments, "favorite lists" and other personal appreciation data.

The emergence of those networks has created a demand for specialized tools, including mechanisms to control abuses and terms-of-use violations. Indeed, the success of social networks has been inevitably accompanied by the emergence of users with non-collaborative behavior, which prevents them from operating evenly. Those behaviors include instigating the anger of other users (*trolling*, in the web jargon), diffusing materials of genre inappropriate for the target community (e.g., diffusing advertisement or pornography in inadequate channels), or manipulating illegitimately popularity ratings (which we call here *ballot stuffing*).

Non-collaborative behavior pollutes the communication channels with unrelated information, and prevents the virtual communities from reaching their original goals of discussion, learning and entertainment. It alienates legitimate users and depreciates the social network value as a whole [1].

In addition to the intricacies inherent to semantic classification, the challenges of content filtering are aggravated by the sheer amount of data social networks host and distribute. An automatic algorithm may be a strong ally to allow the detection of problematic content, but it is crucial that the number of false positives is kept low; otherwise the human agents will be overwhelmed.

In this paper, we address two forms of non-cooperative behavior.

The first is the posting of material considered inappropriate for the audience of the community, like pornographic or violent videos. Some hosts prohibit the posting of this material altogether, while others allow it, provided that it is especially flagged as "adult content". Nevertheless, the content still ends up appearing where it is not welcome, because of either user ignorance of the rules, or full-fledged malice, when it is used to elicit revolted or shocked reactions from other users. Social networks face particular challenges, since content hosts and providers may face economic and, in some jurisdictions, even legal drawbacks (see § 1 in [2]) if they not provide adequate means to protect their users from that kind of material.

The second is the manipulation of video popularity by "ballot stuffing" of video answers. Video answer is a very popular feature of some social networks, like *YouTube*, where the user can post a video in response to another. The most answered videos are featured in a prominent list, creating the incentive for non-cooperative behavior. A user wishing to inflate the popularity of a video, will post many answers (with little content value), sometimes using several different user accounts. This behavior averts the reliability of the recommendation system, irritates other users and consumes a lot of server resources.

We propose a simple scheme, inspired on voting algorithms, a popular technique which has been used in many tasks ranging from parameter estimation [38] to object detection [7] and video classification [39]. Using our scheme, we are able to combine several evidences (coming from classifiers using different features computed over different video elements) in order to obtain the decision on whether or not the video belonging to the unwanted class (e.g., pornography, violence, etc.). The idea is to ask several classifiers which label they attribute to the video. The "opinions" are counted, and the final decision is given by majority vote.

This paper presents two main contributions. First and foremost, the rigorous evaluation of several combinations of descriptors on three applicative scenarios, which has consistently indicated that a representation based on spatiotemporal bags of features is more discriminating than all alternative evidences. This finding is consequential, because the state-of-the-art still relies heavily on static features for video classification — for example, for pornography detection, the overwhelming majority of methods is based on color or texture-based skin detection [].

In addition, we show that, with the right choice of features, a simple scheme, based on majority voting, is able to cope well with the difficult task of content filtering for social networks. We have evaluated our technique in very challenging datasets, conceived to represent the diversity of social networks, obtaining very promising results, with a good compromise between selectivity and specificity.

It is noteworthy that architecture proposed is very flexible, and can being easily adapted to any high-level concept the user might be interested in detecting.

## 2 Prior art

### 2.1 Video feature extraction

Semantic classification of visual documents has only become feasible after the emergence of effective feature extraction algorithms. Many of those may be applied to video, some being just still-image descriptors of individual frames, others being specially conceived to take into account the spatiotemporal nature of the moving image.

Though global image descriptors may be employed to characterize video frames, in the recent years a great deal of interest has been directed to local descriptors. Those are associated to different features of the image (regions, edges or small patches around points of interest) and have been shown to provide great robustness and discriminating power [17][18][24][25][26][27].

The most popular local descriptor, SIFT [7], is both a point of interest detector, based on differences of Gaussians and a local descriptor, based on the orientations of grayscale gradients. Using SIFT, visual content is represented by a set of scale and rotation invariant descriptors, which provides a characterization of local shapes. The generated descriptors allow for adequate levels of affine, viewpoint and illumination invariance.

Since color information is considered important for many tasks (e.g., nude detection), color extensions of SIFT have been proposed [9][21]. For example, a SIFT descriptor adapted to carry hue information (aptly named HueSIFT) had been proposed [9]. It provides color distinctiveness in addition to shape distinctiveness.

Intuition tells us that temporal information should be of prominent importance for recognition tasks in videos, for being likely to indicate interesting patterns of motion. Considering that, a few local features detectors and descriptors have been proposed, take into account the temporal nature of video [6][13][14][15][16]. For example, STIP [6] is designed as a differential operator, simultaneously considering extrema over spatial and temporal scales that correspond to particular patterns of events in specific locations. It extends the Harris corner detector [8] to the temporal domain, finding interest points as moving corner changes direction across a sequence — if the corner movement is constant, no interest point is detected. This allows detecting noteworthy "events" on the video sequence.

## 2.2 Codebooks of visual features

The discriminating power of local descriptors is extremely advantageous when matching objects in scenes, or retrieving specific target documents. However, when considering high-level semantic categories, it quickly becomes an obstacle, since the ability to generalize becomes then essential. A solution to this problem is to quantize the description spaces by using codebooks of local descriptors, in a technique sometimes named "visual dictionary". The visual dictionary is nothing more than a representation which splits the descriptor space into multiple regions, usually by employing non-supervised learning techniques, like clustering. Each region becomes then a "visual word".

The idea is that different regions of the description space will become associated to different semantic concepts, for example, parts of the human body, corners of furniture, vegetation, clear sky, clouds, features of buildings, etc. Yet, it is important to emphasize that this association is latent; there is no need to explicitly attribute meanings to the words (Figure 1). The technique has been employed successfully on several works for retrieval and classification of visual documents [1][5][11].

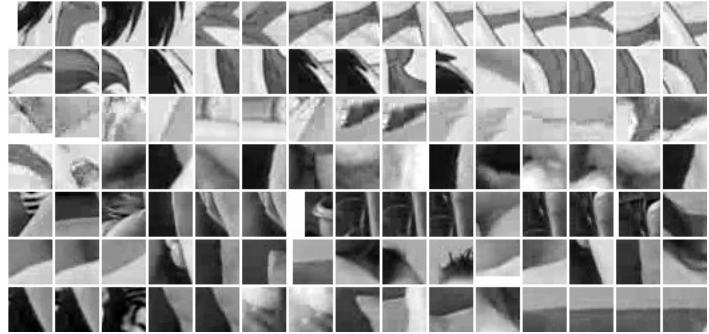

**Figure 1: Image patches associated to descriptors belonging to a typical visual word. There is no explicit association between words and concepts in the visual dictionary technique: the "word" is defined by appearance criteria induced by the descriptor space. Nevertheless we expect the "words" to have some latent semantic charge.**

Once the codebook is obtained, description is greatly simplified, since it is no longer based on the exact value of descriptors, but only on their associated "words". The condensed description may be, for example, a histogram or simply a set of the words the video contains. This has two advantages: the rougher description is better adapted to complex semantics; and the computational burthen is alleviated, since algorithms now operate on a single summarized description, instead of a myriad of individual local descriptors.

Building the dictionary requires the quantization of the description space, which can be obtained by a clustering algorithm. However, state-of-the-art clustering methods are seldom (if ever) conceived for the needs of visual dictionary construction: high-dimensional spaces, large datasets and a large number of clusters. The commonest choice found in the literature is a combination of aggressive sub-sampling of the dataset, dimensionality reduction using PCA (*Principal Component Analysis*), and clustering using a simple or hierarchical *k-means* algorithm with Euclidean distance. This typical choice however, may be considerable faulty on several grounds [22], and the design of good methods for visual dictionary creation is an active, open area of inquire.

In addition to moderating the discriminating power of descriptors, the dictionaries allow adapting to visual documents techniques formerly available only to textual data. Among those borrowings, one of the most successful has been the technique of *bags of words* (which considers textual documents simply as sets of words, ignoring any inherent structure). The equivalent in the CBIR universe has been called *bags of visual words, bags of features* or *bags of visual features,* sometimes abbreviated as BoVF. It greatly simplifies document description, which becomes a histogram of the visual words it contains (Figure 2). The introduction of this technique had a huge impact on content-based retrieval and classification of visual documents [12].

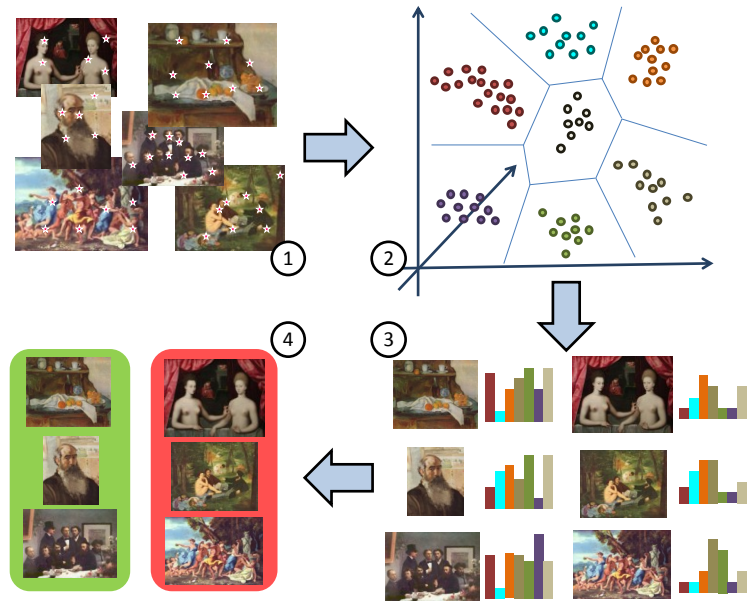

**Figure 2: Content-based classification using BoVF (bags of visual features) — The processing steps are (1) extraction of the local descriptors; (2) quantification of the descriptor space to create the visual dictionary; (3) summarized description of the visual documents as histograms of visual words; (4) use of the summarized description in the classification.**

The straightforward extension of BoVF to video uses individual frame images (or selected keyframes). This allows representing semantic concepts which are independent from motion. However, previous works in human annotation of video databases [40], indicate that even for humans, many important concepts can only be adequately apprehended by taking into account the temporal aspects of video. Therefore, an interesting possibility is making codebooks of space-time local descriptors [6], which take into account the dynamic aspects of video. In this work we evaluate the performance of both static and "motion-aware" bags of features.

## 2.3 Video content filtering

The importance of pornography detection in visual documents is attested by the large literature on the subject. The vast majority of those works is based on the detection of human skin, and suffers from a high rate of false positives in situations of non-pornographic body exposure (like in sports). Some works use secondary criteria (like the shape of the detected skin areas, rejection of facial close-ups, etc.) to lower this rate. A comprehensive survey on skin-detection based methods may be found in [2].

Few methods have explored other possibilities. Bags of visual features (explained in the previous section) have been employed for many complex visual classification tasks, including pornography detection in images and videos [1][4][10]. Those works, however, have explored only bags of static features. Kim et al. [3] compare the effectiveness of several MPEG-7 features, but again, concentrate only on static features, ignoring those related to motion. Very few works have explored spatiotemporal features or other motion information for detection of pornography [41][42][43]. Jansohn [43] uses bags of static visual features and analysis of motion, including motion histograms, as separate evidences, but does not consider bags of spatiotemporal features

In contrast to pornography detection, the identification of pollution in online video social networks is a new topic with very few published works. After extensive research, we could only find two [2][3]. The first tries to identify non-cooperative *users* and addresses both spamming and ballot stuffing, by analyzing parameters like tags, user profile, the user posting behavior and the user social relations. The second classifies videos, accordingly to their pattern of access by users, into three categories: quality (the normal ones), viral (videos which experience a sudden surge in popularity) and junk (spurious videos, like spam). Neither work use the video content itself for classification.

Violence detection has been addressed in works targeting applications as diverse as surveillance systems and movie rating. As expected, the application scope greatly affects how the problem is attacked: for example, in video-surveillance, movies are often black and white, noisy and silent [32][33]; in feature action movies, the soundtrack is often very indicative of the scene action, and so on [35][36] (to the point that some works are based solely on soundtrack evidences [34]).

In [32], a hierarchical approach for detection of violence in surveillance videos is proposed. Several actions involving two people are detected: fist fighting, kicking, hitting with objects, among others. The information of motion trajectories of the image structures in the scenes is obtained with the computation of acceleration measure vectors and their jerks. However, this method poses some limitations, failing for situations involving more than two people and when fighters fall down to the ground. Siebel and Maybank [33] developed a surveillance system for aiding human operators in monitoring undesirable events in a metro station. In [37], regions whose color indicated the presence of skin and blood are analyzed to detect aggressive actions in movies. Later, motion intensities of those regions of interest are computed, higher values indicating violence. We could not find any of approach dealing within the diversity of videos found in social networks, neither employing bags of visual features representations.

## 3 The proposed scheme

The proposed scheme is very simple and works by extracting elements from the video (shots, frames, keyframes, etc.), extracting features from those elements (global features, bags of visual features based on local features, statistics, etc.) and training different classifiers for each type of feature used. In the classification phase, the classifier opinion is asked for each individual video element, and the final decision is reached by majority voting. The whole scheme is illustrated on Figure 1 and explained, in detail, below.

**Pre-processing (video element and feature extraction) step:**

1. The elements of each video are extracted (shots, frames, keyframes, etc.);
2. The features are extracted from the appropriate elements of the video. Those may be visual features, statistics, etc.

**Training step:**

1. A SVM classifier is created for each type of feature [23]. In our work, we have used a linear kernel (which in preliminary tests, has offered the best results);
2. Each classifier is trained with the corresponding features. Care is taken to balance the classes (positive and negative) so each is given roughly the same number of training samples at this step;

**Classification step:**

1. Each SVM classifier is asked about each single feature concerning all elements of the video related to that feature (i.e., if a feature is computed over keyframes, there will be a feature available for every keyframes, and the corresponding classifier will be asked once for each one of those features);

2. Every time it is enquired, a SVM classifier casts a vote: positive (the video is "unwanted") or negative (the video is "ok");

3. Those votes are counted for all classifiers on all features concerning the video. The majority label is given to the video.

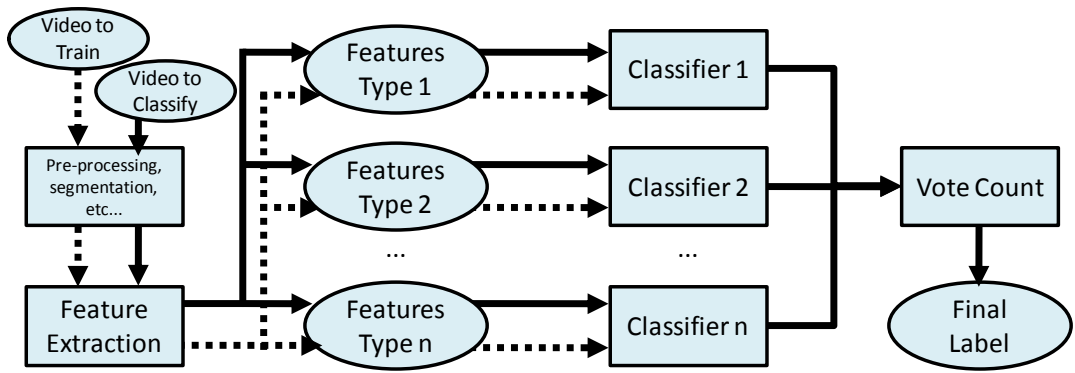

**Figure 3: The proposed scheme for video classification. The data flow for training is represented by the dashed lines, while the data flow for classification is on continuous lines. Each classifier works on a different type of feature (e.g., color histogram, "bag" of local features, etc.) potentially computed over different video elements (frames, shots, etc.). The final label is obtained by majority voting over the opinion of all classifiers. This makes the scheme very robust.**

## 4 First testbed application: pornography detection

Pornography is less straightforward to define than it may seem at first, since it is a high-level semantic category, not easily translatable in terms of simple visual characteristics. Though it certainly relates to nudity, pornography is a different concept: many activities which involve a high degree of body exposure have nothing to do with it. That is why systems based on skin detection [2] often accuse false positives in contexts like beach shots or sports.

A commonly used definition is that pornography is the portrayal of *explicit* sexual matter with the *purpose* of eliciting arousal. This raises several challenges. First and foremost what threshold of explicitness must be crossed for the work to be considered pornographic? Some authors deal with this issue by further dividing the classes [1][3] but this not only fall short of providing a clear cut definition, but also complicates the classification task. The matter of purpose is still more problematic, because it is not an objective property of the document.

**4.1 Test database**

We have opted to keep the evaluation conceptually simple, by assigning only two classes (porn and non-porn). On the other hand, we took great care to make them representative of the diversity found on social networks.

For the pornographic class, we have browsed social networks which only host this kind of material (solving, in a way, the matter of purpose) and sampled 400 videos as broadly as we could (Table I) — the database contains several genres of pornography and depicts actors of many ethnicities (Table II).

For the non-pornographic class we have browsed general-public social networks and selected two samples: 200 videos chosen at random (which we called "easy") and 200 videos selected from textual search queries like "beach", "wrestling", "swimming", which we knew would be particularly challenging for the detector ("difficult").

Figure 4 shows selected frames from a small sample of the database, illustrating the diversity of the pornographic videos and the challenges of the "difficult" non-pornographic ones. Those video clips are readily available at our website[1].

**Table I: A summary of the test database for pornography detection.**

| Class | Videos | Hours | Shots per Video |
|---|---|---|---|
| Porn | 400 | 57 | 15.6 |
| Non-Porn ("Easy") | 200 | 11.5 | 33.8 |
| Non-Porn ("Difficult") | 200 | 8.5 | 17.5 |
| **All videos** | **800** | **77** | **20.6** |

**Table II: Ethnic diversity on the pornographic videos.**

| Ethnicity | % of Videos |
|---|---|
| Asians | 16 % |
| Blacks | 14 % |
| Whites | 46 % |
| Multi-ethnic | 24 % |

---

[1] http://sites.google.com/site/porndetection/

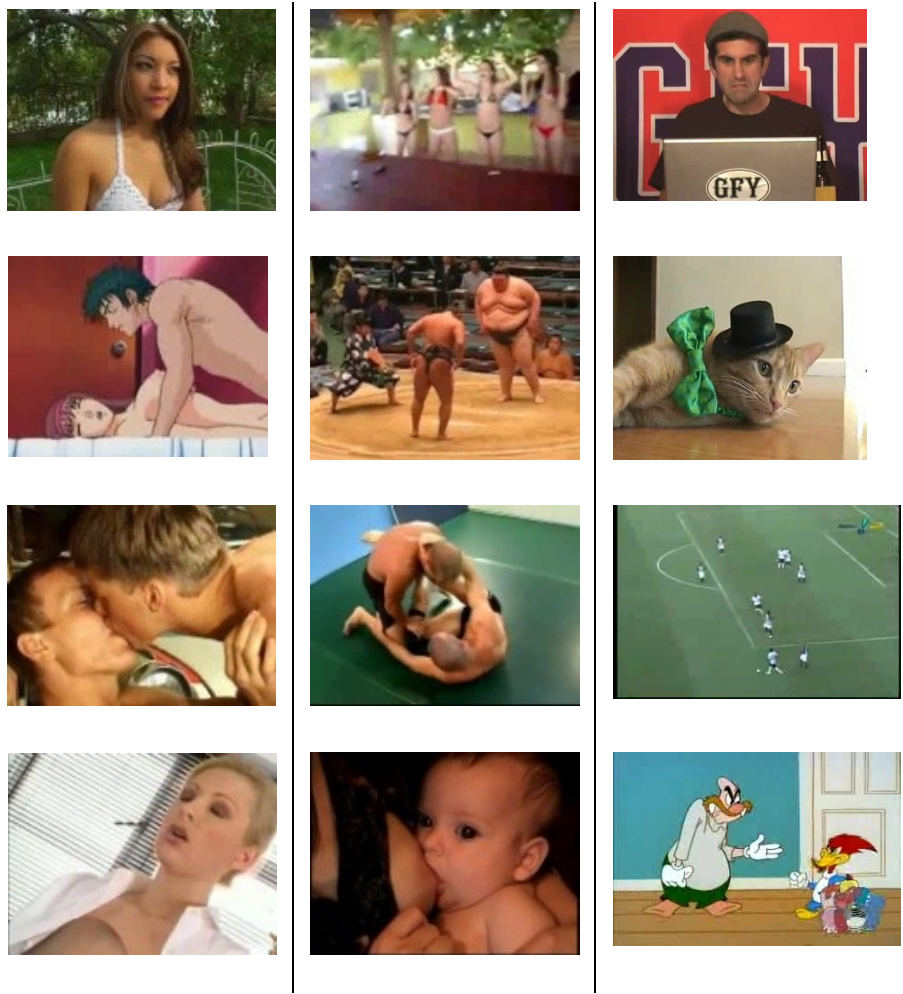

**Figure 4: The challenging task of pornography detection. We have formed our database both to be representative of many forms of pornography (left) and to present a strong challenge to the classifier in terms of the non-porn images (difficult cases in the middle and easy cases at right). A single frame is shown; the entire video-clips are available at our website[1].**

### 4.2 Experimental setup

For this testbed application, the scheme has been parameterized as follows:

1. The two classes considered were porn (positive) and non-porn (negative). It is important to notice that the "easy" and "difficult" non-porn videos are considered in the same classification class. The differentiation is important only for the detailed analysis.

2. The video elements considered are video shots and the middle-frame of each shot. Video shots were obtained by an industry-standard segmentation software[2].

3. The following features computed for the frames:
   - Color Histogram: a normalized 64-bin RGB color histogram;
   - SIFT-BoVF: 5000-bin normalized BoVF using the SIFT descriptor [7];
   - HueSIFT-BoVF: a 5000-bin normalized BoVF using the HueSIFT descriptor [9];

4. The following feature was computed for the shots:
   - STIP-BoVF: a 5000-bin normalized BoVF using the STIP descriptor [6].

Obtaining a baseline to compare with our method was a major challenge since, in general, the numbers reported on the literature are not comparable from one work to another. Often, the databases are given only very cursory description, making next to impossible to make a fair assessment of the actual experimental conditions. Therefore, we have opted to compare ourselves to PornSeer Pro, an industry standard video pornography detection system, which is readily available for evaluation purposes[3]. PornSeer Pro is based on the detection of specific features (like breast, genitals or the act of intercourse) on individual frames. It examines each individual frame of the video.

The experimental design was a classical 5-fold cross-validation, generating approximately 640 videos for training and 160 for testing on each fold.

### 4.3 Results

Figure 5 shows the performance, in the ROC space, of our detector using different combinations of features. It also shows the performance of the baseline method chosen, PornSeer. The graph reveals that several configurations of our detector are not significantly worse than PornSeer, and suggests that a few are significantly better. This latter statement, however, requires a more stringent statistical test, because we are comparing several configurations at once [30][31].

---

[2] http://www.stoik.com/products/svc/
[3] http://www.yangsky.com/products/dshowseer/porndetection/PornSeePro.htm

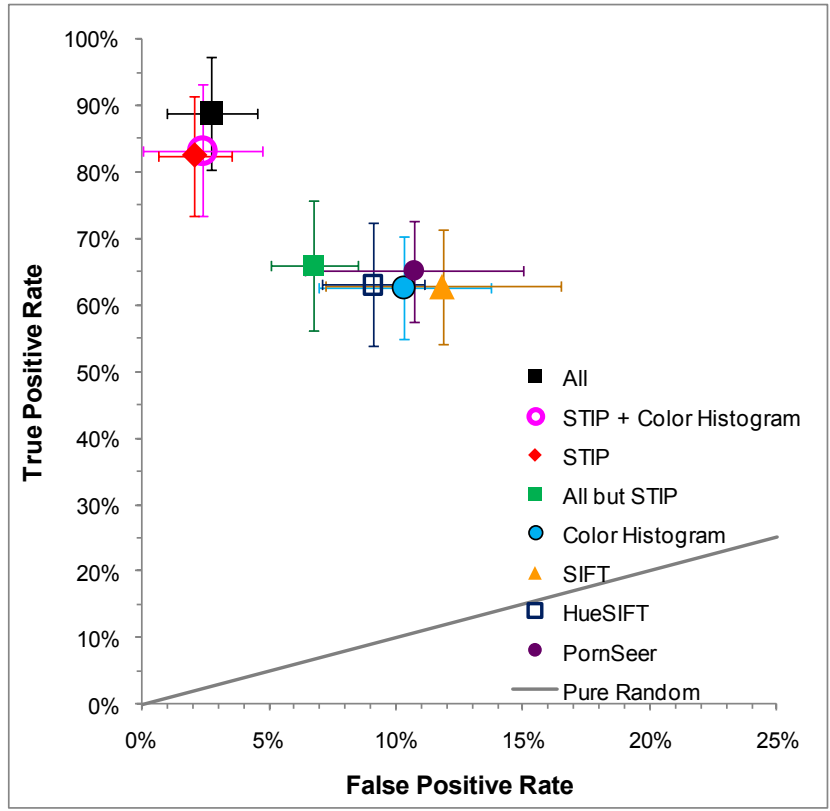

**Figure 5: A few selected points of the above curves. The error bars are confidence intervals on the respective dimensions (for *α* = 0.05). The graph shows that several configurations of our scheme are significantly better than the baseline (PornSeer).**

Therefore, we have performed an ANOVA test, using the 5 runs for all configurations shown in the graph. The model, using the configuration as a factor, was deemed as significant, with a *p*-value of less than 0.01 for both axes. This authorized us to perform pair-wise t-tests between the configurations, Table III shows the *p*-values obtained. We have highlighted values below 0.05 as significant.

**Table III. P-values of pairwise t-test of detector configurations (and PornSeer), with significant differences in boldface. The pairwise tests were done after the ANOVA of all configurations was deemed significant (bottom row).**

| | True Positive Rates | | | | | | | | False Positive Rates | | | | | | | |
|---|---|---|---|---|---|---|---|---|---|---|---|---|---|---|---|---|
| | 1 | 2 | 3 | 4 | 5 | 6 | 7 | 8 | 1 | 2 | 3 | 4 | 5 | 6 | 7 | 8 |
| 1 – Color Hist. | — | .817 | .459 | **.015** | **.049** | .353 | **.041** | .228 | — | .169 | .723 | **.020** | **.006** | .763 | **.021** | .289 |
| 2 – SIFT | .817 | — | .333 | **.008** | **.030** | .249 | **.024** | .327 | .169 | — | .300 | **.001** | **<.001** | .097 | **.001** | **.019** |
| 3 – Hue SIFT | .459 | .333 | — | .075 | .202 | .848 | .174 | .057 | .723 | .300 | — | **.009** | **.002** | .513 | **.009** | .161 |
| 4 – STIP | **.015** | **.008** | .075 | — | .590 | .108 | .650 | **.001** | **.020** | **.001** | **.009** | — | .589 | **.040** | .993 | .178 |
| 5 – STIP + Hist. | **.049** | **.030** | .202 | .590 | — | .275 | .932 | **.003** | **.006** | **<.001** | **.002** | .589 | — | **.012** | .583 | .064 |
| 6 – All but STIP | .353 | .249 | .848 | .108 | .275 | — | .240 | **.038** | .763 | .097 | .513 | **.040** | **.012** | — | **.041** | .444 |
| 7 – All | **.041** | **.024** | .174 | .650 | .932 | .240 | — | **.002** | **.021** | **.001** | **.009** | .993 | .583 | **.041** | — | .181 |
| 8 – PornSeer | .228 | .327 | .057 | **.001** | **.003** | **.038** | **.002** | — | .289 | **.019** | .161 | .178 | .064 | .444 | .181 | — |
| Model *p*-value | | | | .0073 | | | | | | | | .0009 | | | | |

The confusion matrix is another way to express the results shown in the ROC graphs. We showed the matrices for PornSeer (Table IV) and for, arguably, the best configuration of our detector, using just the STIP descriptor (Table V).

**Table IV. The average confusion matrix for PornSeer.**

| | | Video was labeled as | |
|---|---|---|---|
| | | Porn | Non-porn |
| Video was | Porn | 65.1 % | 34.9 % |
| | Non-porn | 12.5 % | 87.5 % |

**Table V. The average confusion matrix for our scheme using STIP.**

| | | Video was labeled as | |
|---|---|---|---|
| | | Porn | Non-porn |
| Video was | Porn | 91.3 % | 8.7 % |
| | Non-porn | 7.5 % | 92.5 % |

### 4.4 Discussion

In its optimal configuration, our scheme is able to correctly identify 9 out of 10 of the pornographic clips, with few false positives. This is very important, since, as we have discussed, the cost of false alarms is high on the social network context, for it tends to overwhelm the human operators. The false positive rate attained may appear high at first, but it must be taken in the context of a very challenging dataset. Considering that half of the non-pornographic test videos were difficult cases, the rates are, actually, low.

It is instructive to study the cases where our method fails. The stubborn false positives correspond to very challenging non-pornographic videos: breastfeeding sequences, sequences of children being bathed, and beach scenes. The method succeeds for many videos with those subjects, but those particular ones have the additional difficulty of having very few shots (typically 1 or 2), giving no allowance for classification errors. PornSeer gave a wrong classification for all those clips.

The analysis of the most difficult false negatives revealed that the method has difficult when the videos are of very poor quality (typical of amateur porn, often uploaded from webcams) or when the clip is only borderline pornographic, with few explicit elements. PornSeer also had difficulty with those clips, misclassifying many of them.

The study of Table III reveals interesting information. It shows that several configurations of our method have significantly better true positive rate than PornSeer, without increasing significantly the false positive rate.

More interestingly, it shows that STIP, the spatiotemporal descriptor, is critical in obtaining those good results. STIP used alone beats, in at least one of the axis, all configurations which do not use STIP; and it ties with all configurations which use STIP in combination with others descriptors. That suggests that not only spatiotemporal information is better for pornography detection in video, but also that combining it with other information (including color!) does not ameliorate the results.

Accumulation of evidences, however, seems to be useful when none of them is much compelling. Though none of the configurations using a single descriptors (other than STIP), is able to beat PornSeer, when the three "weak" descriptors (SIFT, HueSIFT and Color Histogram) are used together, they achieve significantly better detection rates.

# 5 Second testbed application: "ballot stuffing" in video answers

As we have mentioned, a popular feature of *YouTube* is the possibility of answering a video with another, creating "visual threads" of discussion. This makes the system more interactive and provides an additional way for users to find related content. It also provides one of the measures of video popularity: in addition to the most watched, highest ranked and most commented lists, *YouTube* keeps a front page for the most answered videos.

Unfortunately, this reputation mechanism stimulates some users to cheat the system, by creating a large number of spurious answers, in order to appear in the most answered list. Those answers are often very short, very repetitive and have little to none useful content. Sometimes they consist of repetitions of a single frame, rendered in different colors or font styles.

Figure 6 shows some frames extracted from sample videos of a single thread of video-responses. The original video is related to a "worldwide hug campaign". It has been answered with many legitimately related videos, but also with several repetitive posts intended only at inflating its popularity.

It is perhaps useful to contrast those "ballot stuffing" videos with another form of spurious content, which is spam. In a way one is the converse of the other: the "stuffing" video is a content-less video intended to promote the original post; the "spam" is a thematically unrelated video (usually, with non-trivial contents) that uses the popularity of the original to promote itself. "Spamming" is context sensitive — a football video in a culinary thread may be considered spam, the exact same video in a sports thread would be considered legitimate. Detecting spam (as defined here) is a much harder task than detecting stuffing, because one has to take into account the semantic contents of the answer and of the original video. Here we will focus on the problem of stuffing.

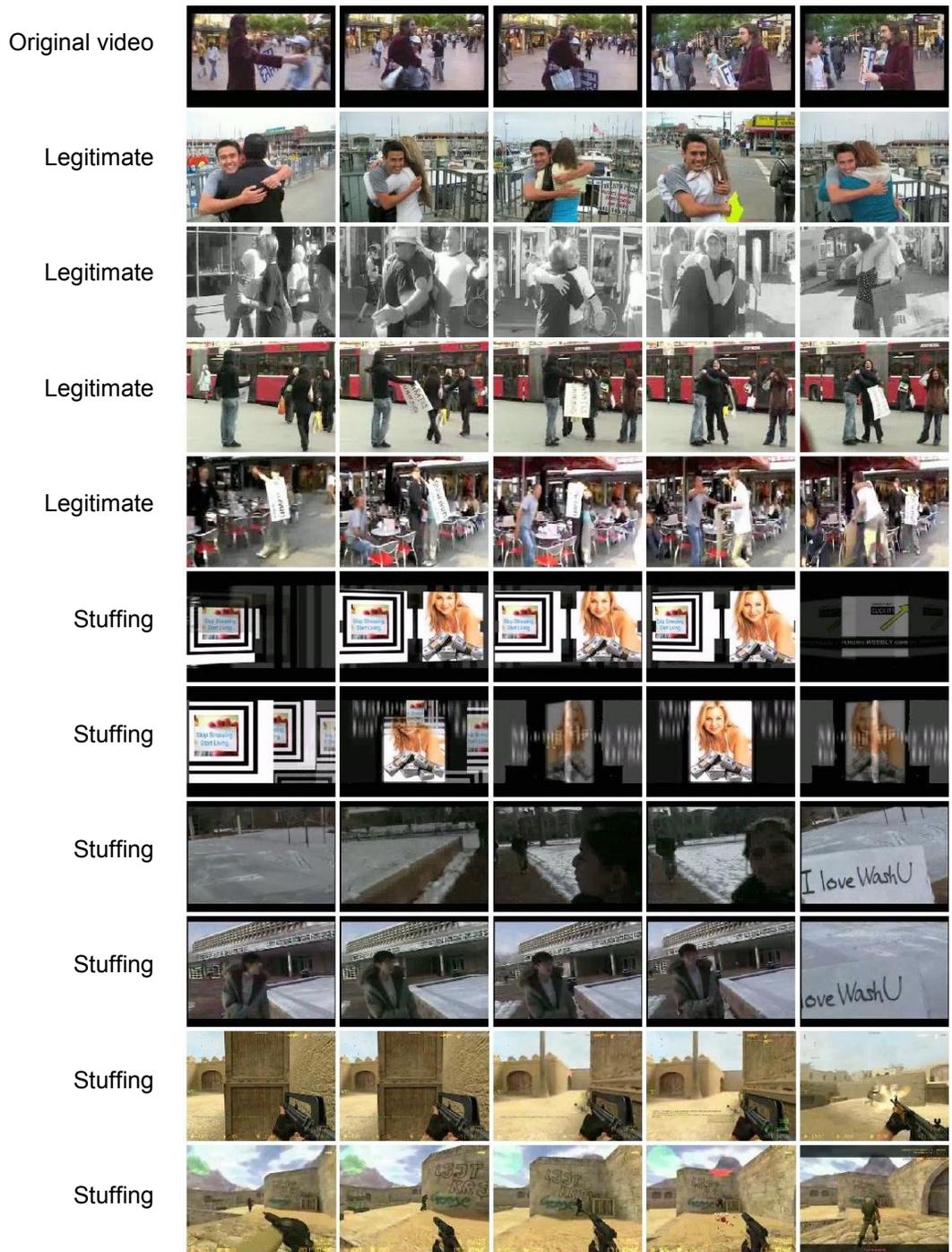

**Figure 6:** Frames extracted from samples of a single video-response thread. The topmost "original" is related to a "worldwide hug campaign". The thread has been answered by several videos, some within the intended thematic ("legitimate"), and some intended only to inflate the popularity of the original video ("stuffing").

### 5.1 Test database

Given the novelty of this application, it is unsurprising that no standard database is available for evaluation purposes. Therefore, we have collected ourselves 8 182 videos from 84 threads, chosen at random from the "Most Responded Videos" list generated by *YouTube*. In addition to the 84 original videos, manual inspection determined 3 420 answers to be "legitimate" and 4 678 to be "stuffing". For the experiment reported here, we have randomly selected 1 000 videos: 500 "legitimate" and 500 "stuffing".

Deciding which videos consist in "stuffing" behavior is not always easy, even for humans. We have decided to consider videos "stuffing" when the same thread contained at least two very similar videos either with very little content or with content unrelated to the original video.

### 5.2 Experimental setup

In this experiment, the evidence fusion scheme has been parameterized as follows:

1. The two classes considered were stuffing (positive) and legitimate (negative).
2. The video elements considered were the entire video and keyframes extracted using a state-of-the art static video summarization method [19].
3. These features were extracted for each keyframe:

   - Global Features:
     - Color Histogram: a normalized 256-bin RGB color histogram;
     - Hue Histogram: a normalized 256-bin histogram of the hue component of color;
     - Zernike Moments: the 10 first Zernike moments;
   - Local Features:
     - SIFT-BoVF: a 5000-bin normalized BoVF using the SIFT descriptor [7];
     - PCA-SIFT-BoVF: a 5000-bin normalized BoVF using the PCA-SIFT descriptor [20].

4. These features were computed for the entire video:

    - Local Features:

        o STIP-BoVF: a 5000-bin normalized BoVF using the STIP descriptor [6];

    - Statistics extracted by the video summarization method:

        o The number of keyframes in the video;

        o The ratio between the number of extracted keyframes and the original number of frames.

The keyframe statistics (used in step 4) are intended as a rough measure of the complexity of the video.

Since the approaches found in literature [28][29] are not comparable to ours — one is concerned with the classification of users; the other does not address the concept of ballot stuffing. We have instead evaluated how different feature choices affected the result.

The experimental design was a classical 5-fold cross-validation, generating approximately 800 videos for training and 200 for testing on each fold.

## 5.3 Results

Figure 7 shows the performance, in the ROC space, of different configurations of our detector, using evidence from different combinations of descriptors. The confidence intervals suggest which differences are statistically significant, but this must be taken carefully, because multiple instances comparison requires stricter statistical tests [30][31].

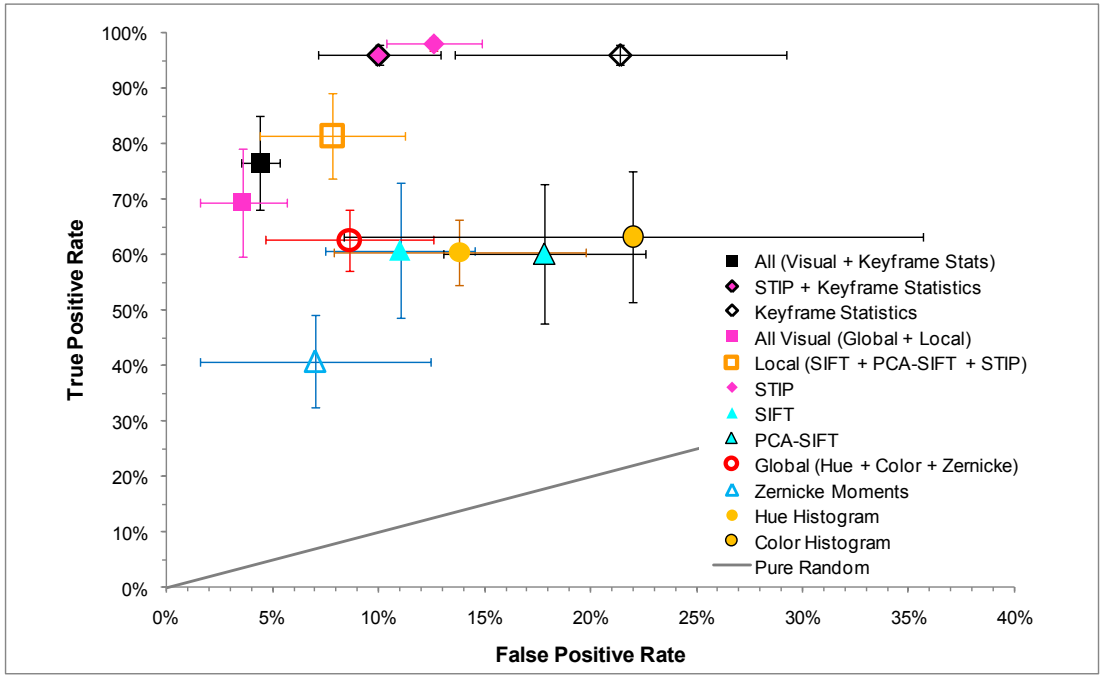

**Figure 7: The performance of the detector in the ROC space, using different features combinations. The error bars are confidence intervals (for α = 0.05).**

Again, we have performed an ANOVA test, using the 5 runs for all configurations shown in the graph. The model, using the configuration as a factor, was deemed as significant, with a *p*-value of less than 0.001 for both dimensions. This authorized us to perform pair-wise t-tests between the configurations, the achieved *p*-values are showed in Table VI (with those above 0.05 highlighted).

**Table VI Pair-wise t-tests of detector configurations. The pair-wise tests were done after the ANOVA of all configurations was deemed significant (bottom row). The circles indicate significative (·<0.001, •<0.01, ○<0.05) and the crosses non-significative (+<0.1, ×≥0.1)p-values.**

| | True Positive Rates | | | | | | | | | | | | False Positive Rates | | | | | | | | | | | |
|---|---|---|---|---|---|---|---|---|---|---|---|---|---|---|---|---|---|---|---|---|---|---|---|---|
| | 1 | 2 | 3 | 4 | 5 | 6 | 7 | 8 | 9 | 10 | 11 | 12 | 1 | 2 | 3 | 4 | 5 | 6 | 7 | 8 | 9 | 10 | 11 | 12 |
| **1 - Hue Hist.** | | × | · | × | × | · | · | × | · | + | • | · | | + | × | × | × | × | + | × | × | ○ | ○ | × |
| **2 - RGB Hist.** | × | | · | × | × | · | · | × | · | × | • | · | + | | • | ○ | × | ○ | × | • | • | · | · | • |
| **3 - Zernicke Mo.** | · | · | | · | · | · | · | · | · | · | · | · | × | • | | × | ○ | × | • | × | × | × | × | × |
| **4 - SIFT** | × | × | · | | × | · | · | × | · | + | • | · | × | ○ | × | | × | × | ○ | × | × | + | × | × |
| **5 - PCA-SIFT** | × | × | · | × | | · | · | × | · | + | • | · | × | × | ○ | × | | × | × | ○ | ○ | • | • | + |
| **6 - STIP** | · | · | · | · | · | | × | · | • | · | · | × | × | ○ | × | × | × | | ○ | × | × | ○ | + | × |
| **7 - Keyfr. Stats** | · | · | · | · | · | × | | · | • | · | · | × | + | × | • | ○ | × | ○ | | • | · | · | · | ○ |
| **8 - All Global** | × | × | · | × | × | · | · | | × | · | · | · | × | • | × | × | ○ | × | • | | × | × | × | × |
| **9 - All Local** | · | · | · | · | · | • | • | × | | ○ | × | • | × | • | × | × | ○ | × | · | × | | × | × | × |
| **10 - All Visual** | + | × | · | + | + | · | · | × | ○ | | · | · | ○ | · | × | + | • | ○ | · | × | × | | × | × |
| **11 - All** | • | • | · | • | • | · | · | · | × | · | | · | ○ | · | × | × | • | + | · | × | × | × | | × |
| **12 - STIP + Stats.** | · | · | · | · | · | × | × | · | • | · | · | | × | • | × | × | + | × | ○ | × | × | × | × | |
| **Model *p*-value** | | | | | **< 0.0001** | | | | | | | | | | | | | **0.0005** | | | | | | |

The results indicate that the best feature configuration, considering the True Positive Rate is the one which uses STIP BoVF alone. Considering also the False Positive Rate, the configurations "All Features", "All Visual" and "All Local" appear statistically tied. The confusion matrix for the configuration using STIP is shown in Table VII, and that for "All Features" is shown in Table VIII.

**Table VII. The average confusion matrix for detector using STIP descriptor.**

|  |  | Video was labeled as | |
|---|---|---|---|
|  |  | Stuffing | Legitimate |
| Video was | Stuffing | **98.0 %** | 2.0 % |
|  | Legitimate | 12.6 % | 87.4 % |

**Table VIII. The average confusion matrix for the detector using "All Features" descriptors.**

|  |  | Video was labeled as | |
|---|---|---|---|
|  |  | Stuffing | Legitimate |
| Video was | Stuffing | 84.0 % | 16.0 % |
|  | Legitimate | **4.4 %** | 95.6 % |

### 5.4 Discussion

Removing content automatically is only possible when false positive rates are very low, because removing a legitimate answer is much more problematic than accepting a spurious one. At this stage, our technique, used in isolation, does not allow such low rates and thus cannot be used to forceful removing content from the social network.

That does not mean, however, that the technique is of no practical value. Two strategies may be employed to make it feasible: one is to combine it with manual inspection of the "suspect" videos, in order to only remove those which are indeed deemed as illegitimate. Another one, more interesting, is to use it to adjust the reputation score of the videos — refusing to count the votes deemed as illegitimate. In this latter scenario, false positives have much less serious consequences, and the "stuffing" behavior is greatly inhibited when users realize that is has become ineffective.

# 6 Third testbed application: violence detection

Violence database suffers from the same problem of pornography detection: although it is a topic of great interest, with an abundant literature, the community lacks a shared violence dataset. In addition, existing works in do not describe the dataset used in enough details to allow a fair comparison. The matter is aggravated by the fact most works have more constrained applicative scope than ours: in the context of video-surveillance or feature movies, the data has more regularity to exploit than in the wild context of social networks, where there is less useful *a priori* information.

## 6.1 Test database

We have assembled a database containing 216 videos, 108 violent and 108 non-violent. The violent video clips come from very diverse contexts, including violent sports, street fights, civil unrest, etc. They also come from diverse sources: broadcasting, cell phones, video-surveillance cameras, etc. Selected frames from the violent class are shown in Figure 8.

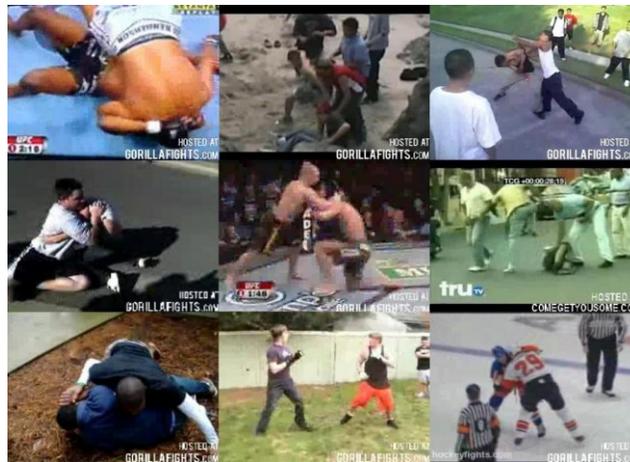

**Figure 8: The violent video dataset. A few situations (in different scenarios) involving aggressive human behaviors are shown: professional fights, street fights, fights in crowds, etc.**

## 6.2 Experimental setup

For this testbed application, the scheme has been parameterized as follows:

1. The two classes considered were violent (positive) and non-violent (negative).

2. The video elements considered are video shots and the middle-frame of each shot. Video shots were obtained by an industry-standard segmentation software[2].

3. The following features computed for the frames:
   - SIFT-BoVF: 100-bin BoVF using the SIFT descriptor [7];

4. The following feature was computed for the shots:
   - STIP-BoVF: a 100-bin BoVF using the STIP descriptor [6].

The evaluation of the classification process was designed and conducted using the traditional 5-fold cross validation scheme, having approximately 160 videos for training and 40 for testing on each fold.

### 6.3 Results

The classification performances are presented in Table IX, for SIFT BoVF, and in Table X for STIP BoVF. Though the results with SIFT are not bad (with over 80% accuracy for both classes) the results using STIP are impressive, scoring perfect results.

**Table IX: Violent video classification using SIFT-BoVF.**

|         |             | Video was labeled as |             |
|---------|-------------|---------|-------------|
|         |             | Violent | Non-violent |
| Video   | Violent     | 80.9%   | 19.1%       |
| was     | Non-violent | 5.0%    | 95.0%       |

**Table X: Violent video classification using STIP-BoVF.**

|         |             | Video was labeled as |             |
|---------|-------------|---------|-------------|
|         |             | Violent | Non-violent |
| Video   | Violent     | 100%    | 0%          |
| was     | Non-violent | 0%      | 100%        |

### 6.4 Discussion

The analysis of the results indicates that local spatiotemporal features are decisive to distinguish between the violent and non-violent descriptors. A closer investigation at the SIFT features indicated that misclassification was, at least to some extent, due to cluttered backgrounds, low-quality frames and scenes of crowded people where the random poses make it extreme challenging to differentiate between violent and non-violent situations. In those situations, the spatiotemporal descriptor (STIP) was able to provide additional motion information, allowing the classifier to reach the right conclusion. In fact, the spatiotemporal events typical of violent videos are so distinctive, that, though the classifier sometimes misses a shot or two, after majority voting is applied, no video is misclassified.

# 7 Conclusions

In all three tasks, the spatiotemporal bags of features performed significantly better than all competing representations. This indicates that motion information is relevant both for identifying a complex semantic category (the pornography and violence tasks) and to measure the relative semantic abundance/scarcity of a video (the stuffing task). This result is not trivial: the state-of-art approaches on pornography detection, for example, are still heavily based upon (static) color and texture skin detection; violence detection approaches often use motion information without considering the advantages of semantic generalization that the codebook representation is able to provide.

When employing that recommended representation, the proposed scheme shows encouraging results in all three tasks, even though the datasets employed were very challenging. A large fraction of unwanted videos is detected, without incurring in excessive false negatives.

In this article, we have evaluated different *visual* features as competing representations for the task of content filtering. Therefore, we have considered visual content as the main source of information. However, it is important to note that the proposed scheme disregards the media modality from where the features are extracted: textual, soundtrack and social interaction information could all be fed to the classifiers. In fact, we believe that, for extremely complex semantic tasks (like spam detection), a multimodal approach is needed to warrant the best performance possible, and we are currently developing our investigations in that direction.

Though our experiments indicate that incorporating information from other descriptors does not significantly improves the performance of the scheme using only STIP, we would like to explore, in the future, non-trivial ways of incorporating this information. We were, for example, surprised by the fact the addition of color information did not improve the results (STIP is "color blind", and, intuitively, color should be an important indicator of pornographic content). We would like to test if incorporating color directly inside the descriptor might improve the results.

One interesting observation for the pornography and violence applications is that label confidence is asymmetric on the training phase. When a training video is labeled as negative, we may be confident that none of its elements is negative. But a training video is labeled as positive may contain several negative shots, frames and keyframes where no violence or pornography is present (opening and closing credits, "cut scenes", etc.). Therefore, at least for the positive class, the training is only weakly supervised. Future versions of the scheme could take this information into account.